\documentclass[11pt]{article}

\usepackage[final]{acl}

\usepackage{times}
\usepackage{latexsym}

\usepackage[T1]{fontenc}

\usepackage[utf8]{inputenc}

\usepackage{microtype}

\usepackage{inconsolata}

\usepackage{graphicx}

\usepackage{booktabs}
\usepackage[whole]{bxcjkjatype}
\usepackage{tabularx}

\title{WikiOriginQA: Automated Construction of a QA Dataset with Knowledge Origins for Analyzing Cultural Bias}

\author{
  Kento Haneda$^{1,2}$,
  Yosuke Kishinami$^{1}$,
  Ryo Fujii$^{1}$,
  Keisuke Sakaguchi$^{2}$,
  Makoto Morishita$^{1}$
\\
  $^{1}$Future Corporation,
  $^{2}$Tohoku University
\\
  \texttt{team-nlp-research@future.co.jp}
}

\usepackage{pifont}
\newcommand{\memo}[1]{}
\newcommand{\wabun}[1]{}

\begin{document}
\maketitle
\begin{abstract}
Existing datasets on cultural knowledge have been typically created by human annotators familiar with each culture.
However, the process requires considerable human effort, presenting a significant challenge to scalability.
To address this challenge, we introduce \textbf{WikiOriginQA}, a QA dataset automatically constructed by objectively representing the origins of knowledge through the creation dates of different Wikipedia language editions.
Experimental results with the created dataset confirm that LLM performance tends to be biased toward Anglophone cultures.
Furthermore, the results suggest the importance of knowledge propagation into the Anglosphere.
These findings demonstrate the validity of our automated approach, enabling cultural bias evaluations with superior scalability across diverse languages without extensive manual annotation.
\end{abstract}

\section{Introduction}
While Large Language Models (LLMs) are being used in many languages, previous studies have pointed out the existence of Anglocentric biases~\cite{wang-etal-2024-countries,hershcovich-etal-2022-challenges}, potentially leading to the exclusion of knowledge from non-English-speaking cultures in NLP technologies.
To understand the extent of cultural bias in multilingual LLMs, it is essential to establish evaluation methods that cover a diverse range of languages and cultural spheres. 
However, many existing evaluation methods require human experts familiar with each culture to manually determine which culture a specific piece of knowledge belongs to~\cite{arora-etal-2025-calmqa, myung2024blendbenchmarkllmseveryday, romanou2024includeevaluatingmultilinguallanguage}, making large-scale expansion to multiple cultures difficult.

\begin{figure}[t]
\centering
\includegraphics[width=0.89\linewidth]{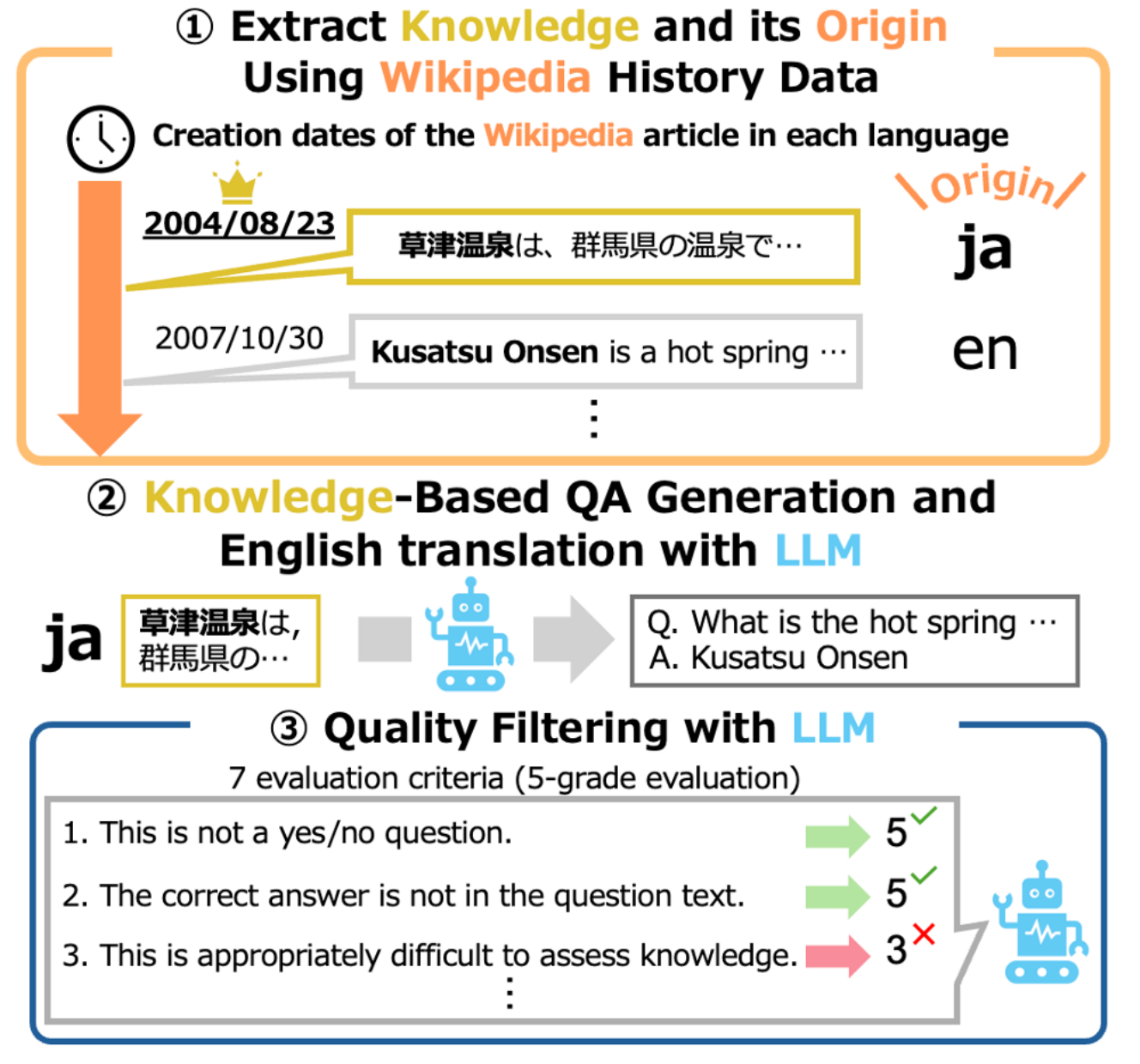}
\caption{The concept of WikiOriginQA. Culture-specific knowledge is expected to appear first in the corresponding Wikipedia language edition.}
\label{fig_concept}
\end{figure}

In this study, we construct a QA dataset for validating cultural bias in a completely automated manner by identifying cultural spheres based on objective information that indicates the origin of knowledge.
Specifically, we refer to the article creation dates across various Wikipedia language editions, based on the hypothesis that knowledge specific to a particular cultural sphere is documented earlier in its associated language than in other languages.
For instance, common knowledge within the Japanese cultural sphere, such as ``草津温泉''~(Kusatsu Onsen, a famous hot spring in Japan), is expected to have its Wikipedia article created in the Japanese version earlier than in any other language~(Figure~\ref{fig_concept}). 
By using the knowledge extracted in this manner, we constructed \textbf{WikiOriginQA},\footnote{\url{https://huggingface.co/datasets/future-architect/WikiOriginQA}} an automatically generated QA dataset for analyzing cultural biases of LLMs.
Our approach achieves high precision in extracting culture-specific knowledge while maintaining high multilingual scalability due to the complete elimination of human intervention.
Experiments using this dataset confirmed that LLMs tend to be biased toward Anglosphere cultures, demonstrating the effectiveness of fully automated datasets for validating cultural bias.

\section{Related Work}
Several benchmarks and datasets have been proposed to evaluate cultural bias in LLMs.
Global MMLU~\cite{singh-etal-2025-global} has confirmed that performance on culture-dependent questions deteriorates in low-resource languages.
\citet{shen-etal-2024-understanding} constructed a benchmark for cultural common sense and demonstrated a decline in model performance regarding non-Western cultures.
Other datasets focusing on cultural knowledge include CaLMQA~\cite{arora-etal-2025-calmqa} for long-form QA, BLEnD~\cite{myung2024blendbenchmarkllmseveryday} for everyday knowledge, and INCLUDE~\cite{romanou2024includeevaluatingmultilinguallanguage} for licensing exams. 
Furthermore, CANDLE~\cite{10.1145/3543507.3583535} efficiently collects cultural knowledge from the English corpora focusing on categories such as food and clothing. These studies have reported Anglocentric biases in LLMs.

However, constructing these datasets typically requires considerable human effort, such as collection and annotation of questions by native speakers, presenting a significant challenge to scalability especially for low-resource languages.
Furthermore, manual judgment raises concerns regarding subjectivity.

To address this, \citet{jiang2025evaluatinglargelanguagemodel} recently proposed KoLasSimpleQA, an objective approach that extracts culture-specific knowledge by identifying items existing only in a specific Wikipedia language edition.
However, this method fails to capture propagated knowledge that has been subsequently added to other language editions.
Therefore, there is a risk that the dataset becomes biased toward excessively niche information.
While our method is also based on Wikipedia, it incorporates temporal information to offer the advantage of being robust against knowledge propagation.

Furthermore, previous studies have reported a close relationship between language and culture regarding the development of Wikipedia articles, and from this perspective, extracting cultural knowledge using article creation timestamps is a rational approach.
For example, articles on topics of high interest to a specific cultural group tend to exist only in the language edition corresponding to that group~\cite{roy-etal-2020-topic}, individuals closely associated with a particular language are more likely to be documented in that language edition~\cite{10.1371/journal.pone.0114825}, and the cultural background of the editors community influences article contents~\cite{10.3389/fphy.2018.00054}.

\section{WikiOriginQA}
\begin{table}[t]
    \centering
    \small
    \tabcolsep 5pt
    \renewcommand{\arraystretch}{1.1}
    \begin{tabular}{lrrr}
    \toprule
    Language & \# Questions & \% En & \% Other \\
    \midrule
    Afrikaans (af) & 81 & 29.6 & 34.6 \\
    Catalan (ca) & 98 & 16.3 & 32.7 \\
    Estonian (et) & 43 & 23.3 & 39.5 \\
    Japanese (ja) & 108 & 17.6 & 30.6 \\
    Slovak (sk) & 45 & 31.1 & 60.0 \\
    Georgian (ka) & 61 & 13.1 & 31.1 \\
    Pashto (ps) & 26 & 11.5 & 15.4 \\
    Ukrainian (uk) & 41 & 7.3 & 26.8 \\
    \midrule
    English (en) & 120 & - & 64.2 \\
    \bottomrule
    \end{tabular}
    \caption{The composition of WikiOriginQA. \% En and \% Other indicate the percentages of questions whose source items have an article in English and in any other language other than the target language, respectively.}
    \label{data_table}
\end{table}

\begin{table}[t]
    \centering
    \small
    \tabcolsep 5pt
    \renewcommand{\arraystretch}{1.1}
    \begin{tabular}{p{7.3cm}}
    \toprule
     \textbf{Question: }Who is the Japanese Western-style painter who was born in Tokushima City, Tokushima Prefecture (Chuo-dori), withdrew together with Asai Chu from the Kobu Art School in 1878 to form the “Juuichikai,” became an art teacher at Osaka Specialized School in 1879, later oversaw the art department at Doshisha from 1895 to 1917, and additionally published all ten volumes of Taisei Futsu Gagakuhon from Garyukan in 1882–1883?\\ \midrule
    \textbf{Answer: }[``Isana Morizumi'', ``守住勇魚'']\\ \bottomrule
    \end{tabular}
    \caption{Example of an automatically generated QA pair for Japanese culture in WikiOriginQA.}
    \label{qa_example_table}
\end{table}

\begin{table*}[t]
    \centering
    \small
    \tabcolsep 5pt
    \renewcommand{\arraystretch}{1.1}
    \begin{tabular}{c|c}
    \toprule
    Language & Randomly selected examples of extracted source knowledge \\ \midrule
    Afrikaans & Swartvlei {\scriptsize (South African saltwater lake)}，Andrew Motjuoadi {\scriptsize (South African painter)}, Tienie Holloway {\scriptsize (South African writer)} \\
    Japanese & 守住勇魚 {\scriptsize (Japanese painter)}，新党さきがけ {\scriptsize (Japanese political party)}, 十六夜日記 {\scriptsize (medieval Japanese travel diary)} \\
    Pashto & \raisebox{-0.2em}{\includegraphics[height=1em]{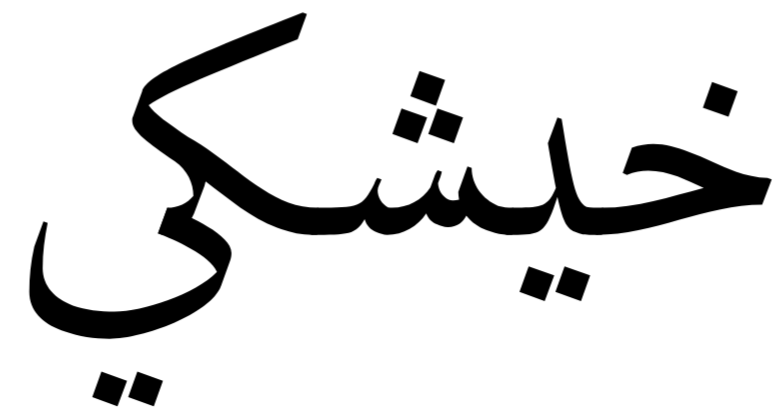}} {\scriptsize (Pashtun tribes)}，\raisebox{-0.2em}{\includegraphics[height=1em]{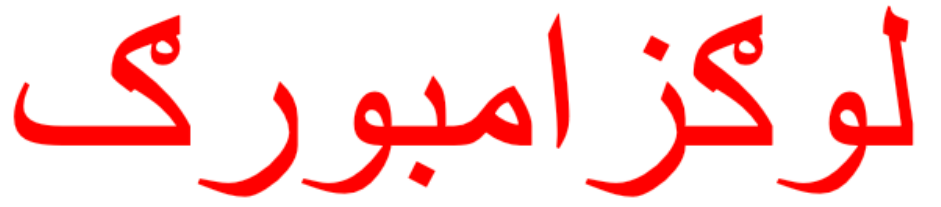}} {\scriptsize (Luxembourg)}，\raisebox{-0.2em}{\includegraphics[height=1em]{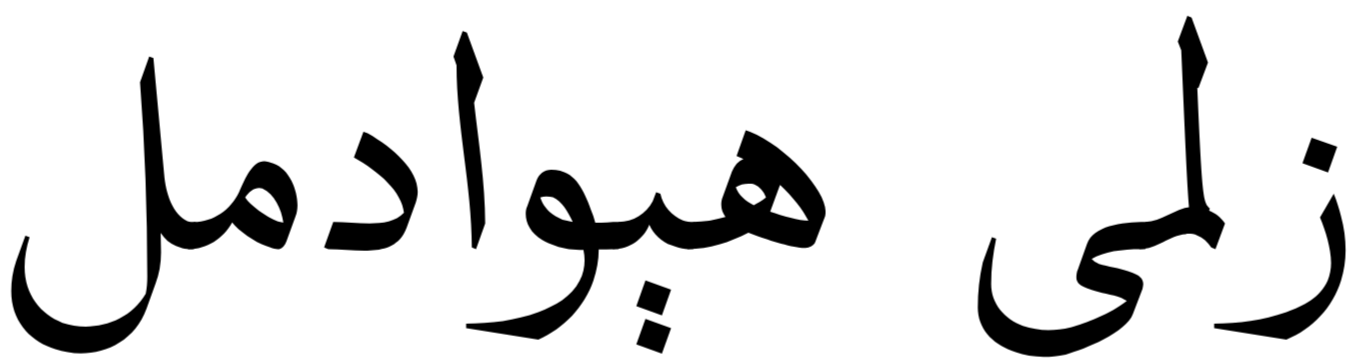}} {\scriptsize (Afghan writer)} \\
    Ukrainian & Пригородок {\scriptsize (City in Ukraine)}，\textcolor{red}{SLC2A8} {\scriptsize (name of a protein)}, Автошлях Т 2601 {\scriptsize (road in Ukraine)} \\
    \bottomrule
    \end{tabular}
    \caption{Randomly selected examples. Red text indicates cases where non-cultural items were incorrectly extracted. Our approach effectively extracted cultural knowledge.}
    \label{q_seed_table}
\end{table*}

In this section, we describe the construction process of WikiOriginQA, a short-answer QA dataset designed to evaluate cultural biases in LLMs.

\subsection{Extraction of Knowledge for QA Data}
To reference the creation dates of Wikipedia articles across different languages representing the same concept, we used Wikipedia dump files and Wikidata. 
Wikidata is a database of items representing topics, concepts, and objects, where a unique entity ID is linked to corresponding Wikipedia articles in each language. 
This ID enables referencing creation dates for all associated Wikipedia language editions.
By comparing creation dates across 342 languages from the List of Wikipedias\footnote{\url{https://meta.wikimedia.org/wiki/List_of_Wikipedias}}, we inferred each item's originating language sphere to serve as the basis for QA data.

\subsection{Generation of QA Data}
\label{subsec:qa-data-construction}
For each extracted entity, we used an LLM to generate a question from its Wikipedia article, with the title as the correct answer.
The prompts were provided in the same language as the corresponding articles. 
In addition, the generated questions were iteratively refined by the LLM to improve their quality.
To minimize the influence of the model's linguistic ability on the results, we translated the QA data into English using an LLM.
Furthermore, we applied an LLM-based filtering to ensure data quality, as illustrated in Figure~\ref{fig_concept}. 
We designed a filter consisting of seven criteria, each evaluated on a 5-point scale, building upon previous studies~\cite{jiang2025evaluatinglargelanguagemodel, he2024chinesesimpleqachinesefactuality, lelkes2021quizstylequestiongenerationnews}.
We retained only questions that received maximum scores across all criteria.\footnote{See Appendix~\ref{sec:details-of-llm-based-filtering} for detailed filtering criteria.} 
We used GPT-5 for these processes with the reasoning effort set to low.

We set gold answers for each question to the original item name supplemented by aliases in Wikidata. 
In addition, to account for instances where a response in the original script is inherently expected (e.g., \textit{kanji} idioms), we accepted answers in the source language.

\subsection{Target Languages and Dataset Details}
\label{subsec:components-of-dataset}
We selected languages whose speakers are concentrated in specific regions as candidates for dataset construction.
Specifically, we selected 75 candidate languages that are designated as official languages\footnote{This includes de-facto official languages.} in exactly one of the 193 United Nations member states.
Among these candidates, we selected 8 languages, listed in Table~\ref{data_table}, considering their regionality and Wikipedia resource availability. 
For each language, we selected 1,000 items and applied the procedure described in Section~\ref{subsec:qa-data-construction} to construct the dataset.
The composition of the dataset is in Table~\ref{data_table}, and an example of questions remaining after filtering is shown in Table~\ref{qa_example_table}.
For comparative evaluation with these languages, we also created the questions in English using the same methodology.

As shown in Table~\ref{q_seed_table}, knowledge specific to each language sphere was effectively extracted.
Furthermore, Table~\ref{data_table} demonstrates ability to capture knowledge propagated to other languages.
Note that there were a small number of cases, such as ``Luxembourg'' in Pashto, where items were extracted separately due to missing Wikidata links, despite existing in other languages.
However, a manual case study of Japanese data confirmed that 96 out of 108 questions (88.9\%) originated from Japan, indicating that such errors are exceptional and demonstrating the high validity of our method.

\section{Experiment}

\begin{figure*}[t]
\centering
\includegraphics[width=0.95\linewidth]{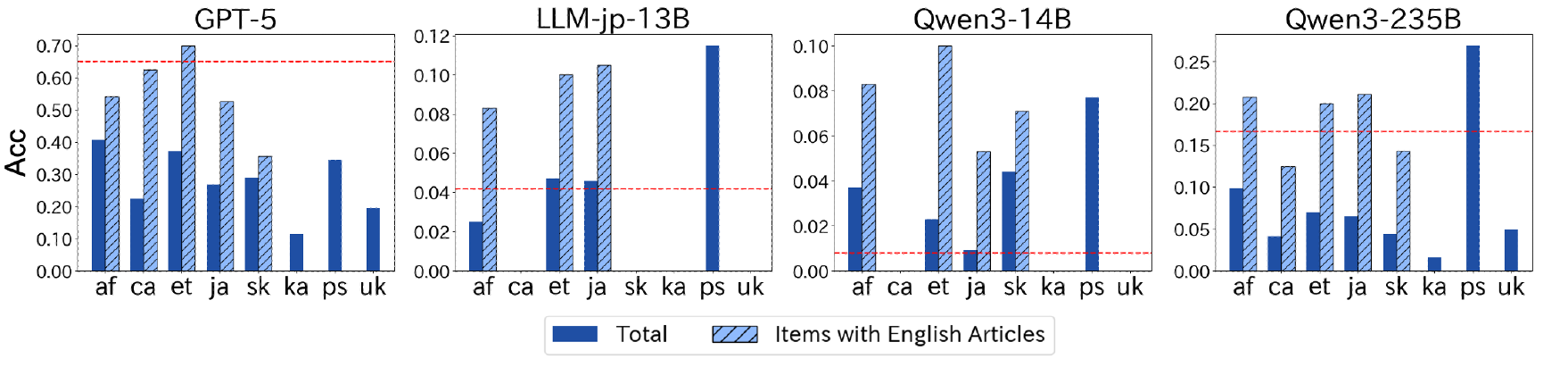}
\caption{Accuracy of questions based on each source language. The red line indicates the accuracy of questions derived from English articles. Higher accuracy was observed for items with English articles. Since the number of questions with English articles was fewer than 10, this analysis was not conducted for ka, ps, and uk.}
\label{each-model-results}
\end{figure*}

\subsection{Settings}
To verify the cultural biases in multilingual LLMs, we prompted several models to generate answers for the questions in WikiOriginQA.
The models used were GPT-5, 
llm-jp-3.1-13b-instruct4~(LLM-jp)~\cite{llmjp-2024-llmjp}, and Qwen3~\cite{yang-etal-2025-qwen3} in various parameter sizes (8B, 14B, 32B, and 235B).
For GPT-5, we set the reasoning effort to low and kept all other parameters to default.
Since the task is in short-answer form, we evaluated correctness not only through Exact Match accuracy~(Acc) but also by using the character-level F1 score~(chrF)~\cite{popovic-2015-chrf}. 

\subsection{Results}
\label{subsec:results}
The accuracy of the models' responses is presented in Figure~\ref{each-model-results}.
Overall, the Acc scores remained at a low level across all languages.
For most models, questions based on English articles achieved higher Acc than those based on other language editions, and this tendency was stronger for models with larger parameter sizes.
For each model, questions whose source items have an English article tended to achieve higher Acc than the overall questions based on that language.\footnote{Detailed results are presented in Appendix~\ref{sec:details-of-results}.}
This indicates that the information propagation to the Anglosphere is essential for a model's knowledge acquisition.

Considering the impact of the translation, we also conducted additional experiments for certain languages using source-language QA pairs without English translation.
As a result, similar to the translated data, higher performance was observed for items that have corresponding English articles.
Furthermore, to address potential extraction errors, we restricted the Japanese dataset to cultural QA pairs that were manually verified, and found that the results mirrored the findings from the overall dataset, which was generated automatically.\footnote{Detailed results are in Appendix~\ref{sec:Results of Original language} and~\ref{sec:Results of manually verified data}, respectively.}

\section{Analysis}
\begin{table}[t]
    \centering
    \small
    \tabcolsep 5pt
    \renewcommand{\arraystretch}{1.1}
    \begin{tabular}{l|rr}
    \toprule
     & GPT-5 & Qwen3-235B \\
    \midrule
    \textbf{True Negative} & 63 / 79 & 93 / 101\\
    \quad Random guess &  44 / 63 &  69 / 93 \\
    \quad Use info in the question & 19 / 63 &  24 / 93 \\
    \textbf{False Negative} & 14 / 79 & 6 / 101 \\
    \quad Spelling variation (English) & 7 / 14 & 2 / 6 \\
    \quad Spelling variation (Other) & 3 / 14 & 0 / 6 \\
    \quad Other (e.g., another name) & 4 /14 & 4 / 6 \\
    \textbf{Other} (e.g., unable to answer) & 2 / 79 & 0 / 101 \\
    \bottomrule
    \end{tabular}
    \caption{Error analysis of GPT-5 and Qwen3-235B. The results demonstrate the validity of automatic evaluation through high true negative rates.}
    \label{error-analysis}
\end{table}

\subsection{Result Analysis}
Experimental results suggest a cultural bias toward the Anglosphere in LLMs, as they achieved higher Acc on questions based on English articles or knowledge propagated into English.
Afrikaans questions were answered relatively accurately; this is likely because English is also an official language in South Africa. 
Consequently, knowledge in Afrikaans is more prone to translation into English compared to other languages, making it easier for the LLM to answer.
However, the intrinsic difficulty of the questions may vary between entities with and without a corresponding English article.
Therefore, further investigation is necessary to distinguish whether the performance gap stems from the availability of English resources for the entity or from the intrinsic difficulty.

\subsection{Error Analysis}
We analyzed the tendencies of the responses that were judged incorrect in the answer of GPT-5 and Qwen3-235B.
Among the 108 questions based on Japanese, 79 for GPT-5 and 101 for Qwen3-235B were judged incorrect by exact match.
Table~\ref{error-analysis} shows the manual verification for these errors.
In most cases, there was no discrepancy between the automatic and manual evaluation, and the responses could reasonably be considered incorrect.

On the other hand, in some cases, responses that should have been judged correct were classified as incorrect.
In many of these cases, as shown in Table~\ref{error-analysis}, the responses were judged incorrect due to factors related to the English translation and the conversion of Japanese, particularly \textit{kanji} into romanized forms.\footnote{A specific example is provided in Appendix~\ref{sec:false-neg}.}
Additionally, some other responses were observed, including GPT-5's refusal to answer. 
Further refinement of question construction and answer evaluation remains future work.

\section{Conclusion}
We proposed a fully automated methodology to construct \textbf{WikiOriginQA}, a highly scalable QA dataset for evaluating cultural biases in LLMs. By identifying knowledge origins using Wikipedia article creation dates, our approach overcomes the scalability and subjectivity issues of traditional human-annotated datasets. Experimental results across various LLMs revealed a significant performance skew toward the Anglosphere, highlighting the persistent risk of excluding non-English cultures. We expect our approach and dataset will help develop more culturally inclusive language models.

\section*{Limitations}

\paragraph{Target Languages}
The experiments conducted in this study primarily aim to demonstrate the practical viability of our proposed dataset construction method, which is inherently scalable to a much broader range of languages. As described in Section~\ref{subsec:components-of-dataset}, there are 75 languages designated as official languages in exactly one country. From this pool, we selected eight languages, taking regional diversity and other factors into account. This selection serves merely as an illustrative example and does not limit the overall scalability of our approach.

However, because our method relies on the intrinsic connection between a language and a specific region, it is difficult to simply apply it to widely spoken languages across multiple regions, such as French or Spanish. While identifying a knowledge's originating region using the IP addresses of Wikipedia editors is a potential alternative approach, we did not adopt it due to concerns regarding long-term sustainability and scalability. Notably, several versions of Wikipedia, including the English edition, have begun phasing out the public display of IP addresses\footnote{\url{https://en.wikipedia.org/wiki/Wikipedia:Temporary_accounts}}.

\paragraph{Coverage in Knowledge Extraction}
Particularly for low-resource languages or those with under-developed Wikipedia editions, it is often the case that cultural items native to that language only exist as English articles or are documented in other major global languages.
In such cases, our method cannot extract knowledge concerning these items, which presents a limitation in achieving high recall when extracting cultural knowledge.
On the other hand, an entity that is documented first in a specific language ahead of all other global languages strongly implies a close connection to that language's cultural sphere, leading to the high-precision extraction of culture-specific knowledge.

\paragraph{Dataset Scale}
The dataset constructed in this study is relatively small in scale owing to financial constraints. Nevertheless, since Wikipedia contains tens of thousands of extractable entities per language, there is substantial room for future expansion.
Specifically, there are over 800,000 items in the Japanese edition, and over 300,000 in the Ukrainian edition, that were created earlier than in any other language version.
In this work, we sampled only a subset of these candidates to build the dataset.

\paragraph{Reliability of Automated Construction}
Because our method focuses on fully automated dataset construction, it is difficult to guarantee the absolute reliability typically associated with human-annotated approaches. However, as demonstrated in Section~\ref{subsec:components-of-dataset}, 88.9\% of the entities in the Japanese subset were confirmed to originate from Japan, indicating that our entity extraction process was successful.
Admittedly, this extraction is not perfectly accurate and leaves room for improvement, however, as demonstrated in Section~\ref{subsec:results}, it does not have a critical impact on evaluating the model's cultural bias, at least regarding the Japanese questions.
Furthermore, regarding the automatically generated QA pairs, although we have not exhaustively verified the entire dataset, the authors manually reviewed a sampled subset and confirmed its high quality. 
Utilizing our method as a foundation while incorporating human-assisted quality control could potentially enable the construction of an even higher-quality dataset; we leave this exploration for future work.

\bibliography{ref}

\appendix
\clearpage
\section{Details of LLM-based Filtering}
\label{sec:details-of-llm-based-filtering}
For data filtering of WikiOriginQA, we employed an LLM to evaluate the questions according to the following criteria:
\begin{enumerate}
    \item The question is not a 'yes'/'no' question.
    \item It is not a question that can only be answered with full sentences like 'how' or 'why.'
    \item The answer does not include unnecessary parts and does not restate parts of the problem.
    \item Questions can be answered using information up until December 31, 2023.
    \item These are questions whose answers are unlikely to change over time.
    \item The questions are not overly simplistic and adequately assess the depth of knowledge of the respondent.
    \item The question stands alone and provides enough context to understand what is being asked without any additional information.
\end{enumerate}

\section{Details of QA Data Generation}
\label{sec:details-of-qa-data-generation}
Given a source such as the Japanese Wikipedia entry for ``草津温泉'' (Kusatsu Onsen), we provide the article text (``草津温泉は，群馬県吾妻郡草津町草津界隈に所在する温泉で...'') to the LLM and instruct it to generate a question with ``草津温泉'' as the answer.
The correct answers initially consist of the entity's title in the original language and its English translation. 
If a corresponding English Wikipedia article exists, its title (in this case, ``Kusatsu Onsen'') is added to the answer set. 
Furthermore, we include aliases listed under Wikidata's \textit{Also known as} property as valid alternative answers. 
For example, the entity "Earth" has Wikidata aliases (``Planet Earth'' and ``The Blue Planet''). We equally treat these as correct answers.

\begin{table*}[t]
    \centering
    \small
    \tabcolsep 5pt
    \renewcommand{\arraystretch}{1.1}
    \begin{tabular}{lc|cccccccccccc}
    \toprule
    & & \multicolumn{2}{c}{GPT-5} & \multicolumn{2}{c}{LLM-jp-13B} & \multicolumn{2}{c}{Qwen3-8B} & \multicolumn{2}{c}{Qwen3-14B}& \multicolumn{2}{c}{Qwen3-32B} & \multicolumn{2}{c}{Qwen3-235B}\\
    Language & \# Questions & Acc & chrF & Acc & chrF & Acc & chrF & Acc & chrF & Acc & chrF & Acc & chrF\\
    \midrule

    Afrikaans (af) &81&.407&.588&.025&.148&.037&.191&.037&.182&.099&.228&.099&.226\\
    Catalan (ca) &98&.224&.460&.000&.148&.010&.165&.000&.170&.061&.272&.041&.261\\
    Estonian (et) &43&.372&.507&.047&.181&.000&.149&.023&.170&.070&.244&.070&.231\\
    Japanese (ja) &108&.269&.403&.046&.131&.009&.155&.009&.167&.056&.224&.065&.229\\
    Slovak (sk) &45&.289&.464&.000&.161&.044&.197&.044&.189&.044&.244&.044&.214\\
    Georgian (ka) &61&.115&.336&.000&.123&.000&.156&.000&.169&.016&.188&.016&.185\\
    Pashto (ps) &26&.346&.581&.115&.255&.154&.272&.077&.240&.269&.452&.269&.438\\
    Ukrainian (uk) &41&.195&.405&.000&.179&.000&.187&.000&.160&.049&.234&.049&.220\\
    \midrule
    
    English (en) &120&.650&.762&.042&.130&.008&.135&.042&.159&.183&.319&.167&.305\\

    \bottomrule
    \end{tabular}
    \caption{Detailed results of the experiment. The table presents the overall evaluation for questions derived from each language. Overall performance was low for both Acc and chrF. However, English-based questions yielded relatively better performance than those from other languages, a tendency that became more prominent as model size increased.}
    \label{appendix_result_table}
\end{table*}

\begin{table}[t]
    \centering
    \small
    \tabcolsep 5pt
    \renewcommand{\arraystretch}{1.1}
    \begin{tabular}{l|cccc}
    \toprule
     & \multicolumn{2}{c}{Translated} &\multicolumn{2}{c}{Original} \\
     & Acc & chrF & Acc & chrF \\
     \midrule
     af (total) & 0.41 & 0.59 & 0.43 & 0.61 \\
     af (w/ en articles) & 0.54 & 0.68 & 0.46 & 0.68 \\
     ja (total) & 0.27 & 0.40 & 0.37 & 0.34 \\
     ja (w/ en articles) & 0.52 & 0.66 & 0.68 & 0.55 \\
    \bottomrule
    \end{tabular}
    \caption{Results using the QA prior to English translation. Trends generally similar to those after English translation were observed.}
    \label{result-of-original-q}
\end{table}

\begin{table}[t]
    \centering
    \small
    \tabcolsep 5pt
    \renewcommand{\arraystretch}{1.1}
    \begin{tabular}{l|cccc}
    \toprule
     & \multicolumn{2}{c}{Total} &\multicolumn{2}{c}{Verified} \\
     & Acc & chrF & Acc & chrF \\
     \midrule
     ja (total) & 0.27 & 0.40 & 0.26 & 0.39 \\
     ja (w/ en articles) & 0.52 & 0.66 & 0.50 & 0.62 \\
    \bottomrule
    \end{tabular}
    \caption{Results of GPT-5's responses on the manually verified data. Trends generally similar to those of the entire dataset were observed.}
    \label{result-of-manually}
\end{table}

\begin{table*}[t]
    \centering
    \small
    \tabcolsep 5pt
    \renewcommand{\arraystretch}{1.2}
    \begin{tabularx}{0.95\textwidth}{l >{\linespread{1.1}\selectfont}X}
    \toprule
    Language & Knowledge \\ \midrule
    Catalan & \textbf{Carlos Santiago Cela Pereira} → A Galician soccer player / \textbf{Neoptòlem de Milet} → A writer who lived in Miletus (a town that once existed in Greece) / \textbf{Vicenç Buron i Llorens} → A historian specializing in the Romanesque of Catalonia / \textbf{Agustí Blanch} → The director of a choir at a church in Barcelona / \textbf{Rectoria de Salitja} → A building in a town called \textit{Salitja} in Catalonia / \textbf{José Antonio Bermúdez}→ A Spanish film sound engineer / \textbf{Na Guardis} → A small island off the coast of Mallorca, Spain / \textbf{Rick Burgett} → An American former professional motocross rider / \textbf{Futur de l'Indicatiu (català)} → A tense expressing the future in Catalan / \textbf{Puig d'en Ponç (Santa Cristina d'Aro)} → A mountain in the Catalonia region \\
    Estonian & \textbf{Ēķini mõis} → A knight's estate in the Livonian region (present-day northeastern Latvia and southern Estonia) / \textbf{Rahvusvaheline Sotsiaalteaduste Rakenduslik Kõrgkool LEX} → A former higher education institution in Tallinn, the capital of Estonia / \textbf{Gertrud Üppis} → An Estonian singer / \textbf{Koigi (Haljala)} → A former town in northern Estonia / \textbf{Ohvriaed} → A garden with religious significance on a farm in southern Estonia / \textbf{Johann Christoph Schmidt} → A clergyman in Estonia and the Livonian region / \textbf{Tõnis Jõgiaas} → An Estonian civil engineer and activist / \textbf{Reiu jõgi} → A river flowing through Estonia and Latvia / \textbf{Heinrich Steding} → The penultimate Master of the Livonian Order / \textbf{Voldemar Horst} → An Estonian sailor \\
    Slovak & \textbf{Bílí Tygři Liberec} → A team in the Czech ice hockey league (the Extralig) / \textbf{Dražkovce} → A town in central Slovakia / \textbf{Xenombrotos z Kosu} → An Olympic equestrian champion from the 5th century BC / \textbf{Olympos (sochár)} → A Greek sculptor from around the 1st century BC / \textbf{Ján Longauer} → A Slovak Catholic priest and educator / \textbf{Devínska hradná s ala} → A natural monument cliff, located at a castle in Slovakia / \textbf{Božský manžel} → An opera written by an Austrian / \textbf{Hronské Kosihy} → A town in southwestern Slovakia / \textbf{1. československá partizánska brigáda J. V. Stalina} → A military force formed during the Slovak National Uprising / \textbf{Rosinka} → A stream in northern Slovakia \\ 
    English & \textbf{Victoria Vesna} → An American artist / \textbf{Huastec people} → The Indigenous people of Mexico / \textbf{Prosaposin} → The name of a protein / \textbf{Canal Street station (New York City Subway)} → A subway station in New York City / \textbf{Liberty Square Blueprint} → An Occupy Wall Street manifesto / \textbf{Infinity (Lee Morgan album)} → An American jazz trumpeter Lee Morgan's album / \textbf{Marianne Francis} → An English evangelical / \textbf{Silas Mix} → An American lawyer / \textbf{Mikey Chung} → A Jamaican musician / \textbf{Butch Gautreaux} → An American politician \\
    \bottomrule
    \end{tabularx}
    \caption{Examples of source knowledge for the QA data across several languages.}
    \label{appendix_qa_base_examples}
\end{table*}

\section{Details of Results}
\label{sec:details-of-results}
Table~\ref{appendix_result_table} shows the detailed QA performance for each source language.

\section{Results of Original Language}
\label{sec:Results of Original language}
Table~\ref{result-of-original-q} shows the results of the GPT-5's responses to each source-language QA prior to English translation.

\section{Results of Manually Verified Data}
\label{sec:Results of manually verified data}
Table~\ref{result-of-manually} presents GPT-5's performance on manually verified Japanese QA.

\section{Examples of Source Knowledge}
\label{sec:examples-of-source-knowledge}
Table~\ref{appendix_qa_base_examples} summarizes randomly sampled examples of the knowledge that served as the basis for the QA data across several languages.

\section{Example of Incorrect Answer by GPT-5}
\label{sec:exmple-of-incorrect-gpt}
\begin{table}[t]
    \centering
    \small
    \tabcolsep 5pt
    \renewcommand{\arraystretch}{1.1}
    \begin{tabular}{p{7.3cm}}
    \toprule
     \textbf{Question: }Who was the person whose childhood name was Tsurunosuke and [$\cdots$] becoming the founder of the Mita \textbf{Nanbu} family?\\ \midrule
    \textbf{GPT-5's Answer: }\textbf{Nanbu} Yukishige \\
    \textbf{Correct Answers: }["Katsunobu Nambu", "南部勝信"] \\ \bottomrule
    \end{tabular}
    \caption{Incorrect answer by GPT-5. The model answers by inferring from the family name \textbf{Nanbu} present in the question text.}
    \label{fail_example_table}
\end{table}

Table~\ref{fail_example_table} shows examples of incorrect responses where the model answered by inferring from the information in the question text.

\section{Examples Erroneously Judged as Incorrect}
\label{sec:false-neg}
\begin{table}[t]
    \centering
    \small
    \tabcolsep 5pt
    \renewcommand{\arraystretch}{1.1}
    \begin{tabular}{p{7.3cm}}
    \toprule
     \textbf{Question: }What is the name of the Class A river in the Arakawa river system [$\cdots$] referred to as ``Nenfutorikawa'' in the Edo-period Shinpen Musashi Fudokikō?\\ \midrule
    \textbf{GPT-5's Answer: }Furo River \\
    \textbf{Correct Answers: }[``Ageless River'', ``不老川''] \\ \bottomrule
    \end{tabular}
    \caption{An example of unintentional incorrect answers．``不老'' is pronounced ``Furo'' and ``川'' means ``river''.}
    \label{false_neg}
\end{table}

Table~\ref{false_neg} shows examples of responses erroneously judged as incorrect.

\section{Target Languages for Retrieving Revision History}
\label{sec:history-language}
In this study, we retrieved the edit histories of 342 active Wikipedia language editions with publicly available data dumps (\url{https://meta.wikimedia.org/wiki/List_of_Wikipedias}) to identify the language of earliest creation for each item.

\section{Filtering Extracted Knowledge Using LLM}
As described in Section~\ref{subsec:components-of-dataset}, knowledge extraction using our method is not perfect. 
To improve extraction accuracy, we used GPT-5 to determine whether the extracted knowledge was cultural and filtered out knowledge that was judged not to be cultural. 
When this filtering was applied to the Japanese data, the extraction accuracy improved from 89\% to 94\%, confirming that LLM-based filtering can contribute to improving extraction accuracy to some extent.

\section{Dataset License}
The WikiOriginQA dataset proposed in this paper is released under the Apache License 2.0 (\url{https://www.apache.org/licenses/LICENSE-2.0}).
\end{document}